\documentclass[11pt]{article}

\usepackage[preprint]{acl}

\usepackage{times}
\usepackage{latexsym}
\usepackage{tikz}
\usepackage{xcolor}
\usetikzlibrary{arrows.meta, positioning}

\definecolor{inkblue}{HTML}{1F3B73}
\definecolor{deepteal}{HTML}{0F6E6E}
\definecolor{plum}{HTML}{6B3E8F}
\definecolor{amber}{HTML}{9C5D0B}
\definecolor{slate}{HTML}{4A4A4A}
\usepackage[T1]{fontenc}

\usepackage[utf8]{inputenc}

\usepackage{microtype}

\usepackage{inconsolata}

\usepackage{graphicx}

\usepackage{listings}
\usepackage{xcolor}
\usepackage{booktabs}

\title{\texttt{pylazaro}: a Python package for anglicism extraction in Spanish}

\author{Elena Álvarez-Mellado \\
  Department of Linguistics \\ Universidad Autónoma de Madrid \\
  \texttt{elena.alvarezm@uam.es} }

\begin{document}
\maketitle
\begin{abstract}
Lexical borrowings are words from one language that are introduced into another language. Identifying lexical borrowings in text is a relevant task for data-centric fields in Linguistics such as lexicography or corpus linguistics, but none of the standard libraries for text processing offers such a functionality. In this paper we present \texttt{pylazaro}, an open-source Python package for the automatic extraction of unassimilated lexical borrowings (mostly anglicisms) from Spanish text. \texttt{pylazaro} offers a single interface to five sequence labeling models that were trained using different libraries, so that users can run and switch between them without having to deal with the idiosyncrasies of each library. We describe the design and usage of the package, contrast the performance of its models with that of general-purpose LLMs (which perform poorly at this task: F1 below 0.40, compared to 0.86 for the best model in \texttt{pylazaro}) and report on its adoption: \texttt{pylazaro} has been downloaded more than 58,000 times and is the library behind \emph{Observatorio L\'azaro}, a resource that monitors anglicism usage in the Spanish press. \texttt{pylazaro} can be installed via PyPI, is documented in \textit{readthedocs} and can be tried through a live demo hosted on HuggingFace Spaces.
\end{abstract}

\section{Introduction}

Lexical borrowings are words from one language that are introduced in another language. 
The process of borrowing is a manifestation of contact between linguistic communities and a prolific source of new words and meanings in a language \citep{haugen_analysis_1950,weinreich_languages_1963,poplack_myths_2012}.
For instance, the French word \textit{weekend} was originally borrowed from English; the word \textit{berde} in Euskara (``green'') was borrowed from Spanish \textit{verde}.

Identifying lexical borrowings in a text is an important task in data-centric fields in Linguistics, such as lexicography, corpus linguistics or historical linguistics. Traditionally, this work has been done by hand, with experts manually annotating corpora and looking up words. 
Providing experts with a tool that can assist them at identifying lexical borrowings in text would automatize a task that is otherwise time-consuming.
However, none of the standard libraries for text processing and annotation (such as \texttt{spaCy}, \texttt{stanza}, etc.) offers such a functionality, nor does there exist (to the best of our knowledge) a publicly-available library specifically devoted to identifying borrowings in running text.  

In this paper we introduce such a tool, \texttt{pylazaro}, a Python package for the task of extracting unassimilated (that is, not yet integrated into the recipient language) lexical borrowings from Spanish text.
\texttt{pylazaro} is available via PyPI\footnote{\url{https://pypi.org/project/pylazaro/}}, its code is released under the MIT license on GitHub\footnote{\url{https://github.com/lirondos/pylazaro}}, it is documented in \emph{readthedocs}\footnote{\url{https://pylazaro.readthedocs.io/en/latest/}}, and a live demo of the library can be freely accessed via HuggingFace Spaces\footnote{\url{https://huggingface.co/spaces/lirondos/pylazaro-demo}} with no installation required.

The contributions of this paper are the following: (1) we describe the design of \texttt{pylazaro}, a package that offers a unified interface to five models for borrowing detection in Spanish that rely on different libraries (Section~\ref{sec:pylazaro}); (2) we present a live web demo that allows users with no programming background to run and compare the models (Section~\ref{sec:demo}); (3) we contrast the performance of the models behind \texttt{pylazaro} with that of general-purpose LLMs on the same task (Section~\ref{sec:llm}); and (4) we report on the adoption of the library and discuss its use cases (Section~\ref{sec:adoption}).

\section{Previous work}
Anglicism extraction is the task of retrieving English lexical borrowings (or \emph{anglicisms}) from non-English texts.
Anglicisms can be single-item (\textit{app}) or multiword (\textit{machine learning}, \textit{fake news}). 
The task of automatically retrieving lexical borrowings from text has proven useful for the preprocessing of linguistic corpora in various languages  \citep{furiassi_retrieval_2007,andersen_semi-automatic_2012,losnegaard_data-driven_2012,serigos_applying_2017} and has previously been framed as a sequence labeling task \citep{alvarez-mellado_overview_2021}, in which relevant in-context spans of text are retrieved from sentences.

In the case of Spanish, the automatic detection of anglicisms has been approached through dictionary lookup and rule-based methods \citep{serigos_applying_2017}, as well as through machine learning models trained on annotated corpora of Spanish newspaper text \citep{alvarez-mellado-2020-annotated,alvarez-mellado_detecting_2022}. The ADoBo shared task \citep{alvarez-mellado_overview_2021} framed the detection of unassimilated borrowings in the Spanish press as a sequence labeling task and attracted systems based on CRFs, BiLSTMs and Transformers \citep{de_la_rosa_adobo_2021,jiang_bert4ever_2021}. However, the models developed in this line of work were released as research code, and using them requires familiarity with the specific libraries in which they were implemented.

There are a few recent libraries aimed at quantitative
tasks in historical linguistics that include identification of borrowings based on etymological data and through the identification of similar words that cannot be diachronically explained as cognates, such as \texttt{LingPy}\footnote{\url{https://lingpy.org/}} \citep{list_lingpy_2021}, \texttt{PyBor}\footnote{\url{https://github.com/lingpy/pybor}} \citep{miller_pybor_2020} or \texttt{LoanPy}\footnote{\url{https://github.com/LoanDB/loanpy}} \citep{martinovic_loanpydatahubloanpy_2023}.
These resources, however, serve a very different purpose from ours, as their aim is not to annotate novel lexical borrowings in a text, but to establish phylogenetic relations between the lexicon of two languages, attest language contact and reconstruct ancient protolanguages or proto-words through cognate identification.  

\section{\texttt{pylazaro}}\label{sec:pylazaro}
\subsection{Overview}

\texttt{pylazaro} is a Python package that takes a text in Spanish as input and returns the lexical borrowings that are present in the text. 

Under the hood, \texttt{pylazaro} is a Python wrapper over the \texttt{Transformers} library\footnote{\url{https://github.com/huggingface/transformers}} and \texttt{flair}\footnote{\url{https://github.com/flairnlp/flair}}. It also relies on \texttt{spaCy}'s\footnote{\url{https://spacy.io/}} \texttt{Token} and \texttt{Span} utilities as building blocks.

\texttt{pylazaro} can be run with five different types of models:
\begin{itemize}
    \item A BiLSTM-CRF model fed with subword embeddings and lexical embeddings pretrained on codeswitching data (this is the best performing model, and the default model used by \texttt{pylazaro}).
    \item A BiLSTM-CRF model fed with subword embeddings and bilingual Transformer-based Spanish-English lexical embeddings.
    \item A Transformer model based on multilingual BERT.
    \item A Transformer model based on Spanish model BETO.
    \item A Conditional Random Field model with handcrafted features.   
\end{itemize}

These five models were introduced and published in prior work \citep{alvarez-mellado_detecting_2022} and each of them exhibits different strengths and weaknesses, some being better at retrieving borrowings in certain sentence positions, with others excelling at different borrowing contexts or shapes \citep{alvarez2024characterizing}. This means that a user may want to switch between models at some point or compare the output produced by two of them. However,  these models were trained using different infrastructures and therefore use different libraries. The point of \texttt{pylazaro} is to offer a single interface that allows for running current (and future) models for anglicism identification using a single entry point and that facilitates switching between models smoothly without the hassle of having to deal with the idiosyncrasies of each of the libraries.

\subsection{How to use \texttt{pylazaro}}
\texttt{pylazaro} can be installed via PyPI and its documentation lives in \textit{readthedocs}.
The user creates a tagger (an object of type \texttt{Lazaro}), which ingests text in Spanish. The tagger will return the lexical borrowings found in the text, allowing for multiple output formats: tuples, dictionary, list of tagged tokens, etc. (see Listing \ref{lst:example}).

The tagger frames the task of identifying borrowings as a sequence labeling task. Therefore, it assigns a label to each token in the sentence following BIO encoding \citep{ramshaw-marcus-1995-text}. The output can then be displayed as a sequence of BIO-tagged tokens (with most of the tokens being labeled as \texttt{O}), or as a series of retrieved spans identified by start and end positions. 

\texttt{pylazaro} retrieves lexical borrowings in general, in other words, it identifies words that come from a language other than Spanish and that have not yet been assimilated into Spanish. Although in theory \texttt{pylazaro} identifies borrowings from any language, in reality the models have been mostly optimized to identify borrowings of English origin. In consequence, the tagger assigns two possible labels: \texttt{ENG} for spans (or tokens) labeled as being of English origin, and \texttt{OTHER} for borrowings of any other language (such as Japanese, French, etc.).

\begin{lstlisting}[caption={Detecting borrowings with pylazaro.}, label={lst:example}]
from pylazaro import Lazaro
tagger = Lazaro()
text = "Fue un look sencillo. Se celebra un festival de 'anime'."
output = tagger.analyze(text)
output.borrowings_to_tuple()
[('look', 'en'), ('anime', 'other')]
output.anglicisms_to_tuple()
[('look', 'en')]
output.other_to_tuple()
[('anime', 'other')]
output.borrowings_to_dict()
[{'borrowing': 'look', 'language': 'en', 'start_pos': 2, 'end_pos': 3}, {'borrowing': 'anime', 'language': 'other', 'start_pos': 11, 'end_pos': 12}]
output.anglicisms_to_dict()
[{'borrowing': 'look', 'language': 'en', 'start_pos': 2, 'end_pos': 3}]
output.other_to_dict()
[{'borrowing': 'anime', 'language': 'other', 'start_pos': 11, 'end_pos': 12}]
output.tag_per_token()
[('Fue', 'O'), ('un', 'O'), ('look', 'B-ENG'), ('sencillo', 'O'), ('.', 'O'), ('Se', 'O'), ('celebra', 'O'), ('un', 'O'), ('festival', 'O'), ('de', 'O'), ("'", 'O'), ('anime', 'B-OTHER'), ("'", 'O'), ('.', 'O')]
\end{lstlisting}

\subsection{Selecting a model}

By default, \texttt{pylazaro} loads the best performing model (the BiLSTM-CRF model with codeswitch embeddings). Users can select any of the other models when creating the tagger by specifying the type of model and the model file (see Listing~\ref{lst:models}). Regardless of the model selected, the tagger is used in exactly the same way and returns the same output formats, which means that switching between models requires changing a single line of code. The models are hosted on the HuggingFace Hub and are downloaded automatically the first time they are used\footnote{The CRF model requires installing  \texttt{pylazaro} with the extended installation (instead of the default installation), which requires additional dependencies.}. 
\begin{lstlisting}[caption={Selecting different models in \texttt{pylazaro}.}, label={lst:models}]
from pylazaro import Lazaro
# BiLSTM-CRF with BETO/BERT embeddings
tagger = Lazaro(model_type="bilstm",
    model_file="lirondos/anglicisms-spanish-flair-bert-beto")
# Transformer model based on mBERT
tagger = Lazaro(model_type="transformers",
    model_file="lirondos/anglicisms-spanish-mbert")
\end{lstlisting}

\subsection{Architecture}

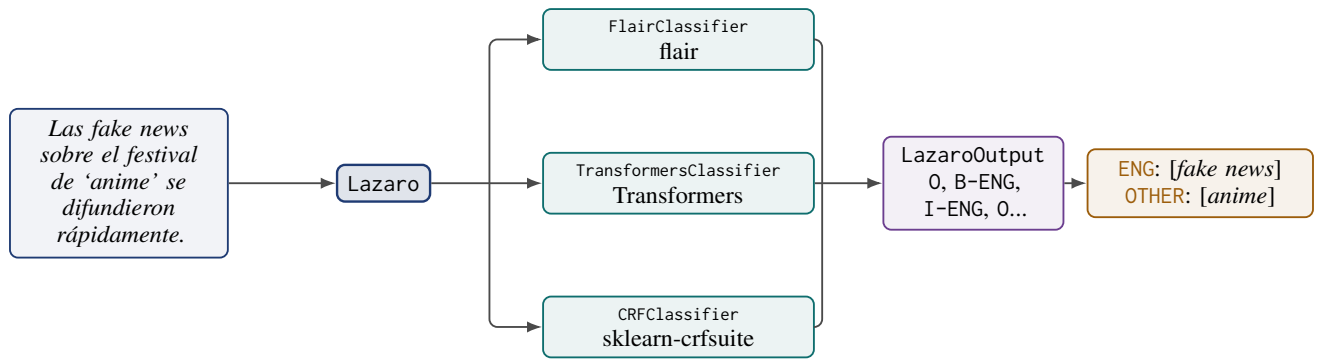
\begin{figure*}[t]
\centering
\begin{tikzpicture}[
  font=\footnotesize,
  io/.style={draw=inkblue, line width=0.7pt, rounded corners=3pt, fill=inkblue!6,
             align=center, inner sep=4pt, text width=2.6cm},
  core/.style={draw=inkblue, line width=0.9pt, rounded corners=3pt, fill=inkblue!14,
               align=center, inner sep=4pt},
   clf/.style={draw=deepteal, line width=0.7pt, rounded corners=3pt, fill=deepteal!8,
              align=center, inner sep=4pt, text width=3.3cm},
  res/.style={draw=plum, line width=0.7pt, rounded corners=3pt, fill=plum!8,
              align=center, inner sep=4pt, text width=2.1cm},
    outbox/.style={draw=amber, line width=0.7pt, rounded corners=3pt, fill=amber!8,
              align=center, inner sep=4pt, text width=2.7cm},
  arr/.style={-{Latex[length=2mm]}, draw=slate, line width=0.7pt, rounded corners=3pt},
  link/.style={draw=slate, line width=0.7pt, rounded corners=3pt}
]

\node[io]   (input)  at (0,0)    {\textit{Las fake news sobre el festival de `anime' se difundieron r\'apidamente.}};
\node[core] (lazaro) at (3.5,0)  {\texttt{Lazaro}};
\node[res]  (result) at (11.3,0) {\texttt{LazaroOutput} \\  \texttt{O}, \texttt{B-ENG}, \texttt{I-ENG}, \texttt{O}...};
\node[outbox] (output) at (14.3,0) {\textcolor{amber}{\texttt{ENG}}:  [\textit{fake news}]\\ \textcolor{amber}{\texttt{OTHER}}: [\textit{anime}] };

\node[clf]  (trans)  at (7.4,0)  {{\scriptsize\texttt{TransformersClassifier}} \\ Transformers};
\node[clf] (flair) at (7.4,1.9)  {{\scriptsize\texttt{FlairClassifier}} \\ flair};
\node[clf] (crf)   at (7.4,-1.9) {{\scriptsize\texttt{CRFClassifier}} \\ sklearn-crfsuite};

\coordinate (bin)  at (4.9,0);
\coordinate (bout) at (9.3,0);

\draw[arr] (input) -- (lazaro);
\draw[link] (lazaro.east) -- (bin);
\draw[arr]  (bin) |- (flair.west);
\draw[arr]  (bin) -- (trans.west);
\draw[arr]  (bin) |- (crf.west);
\draw[link] (flair.east) -| (bout);
\draw[link] (trans.east) -- (bout);
\draw[link] (crf.east)   -| (bout);
\draw[arr]  (bout) -- (result.west);
\draw[arr] (result) -- (output);

\end{tikzpicture}
\caption{Architecture of \texttt{pylazaro}. The \texttt{Lazaro} class receives the input text and dispatches it to the classifier that corresponds to the model selected by the user. Each classifier loads its model using the library it was trained with and returns a sequence of BIO tags, which are wrapped in a common output object that implements the different output formats. Adding a new model only requires implementing a new classifier.}
\label{fig:architecture}
\end{figure*}

Internally, \texttt{pylazaro} separates the user-facing interface from the models that perform the prediction (see Figure \ref{fig:architecture}). The \texttt{Lazaro} class acts as the single entry point: it validates the parameters provided by the user and, depending on the type of model requested, instantiates the corresponding classifier (e.g. \texttt{FlairClassifier} for the BiLSTM-CRF models). Each classifier is in charge of loading its model using the library it was trained with, producing a sequence of BIO tags for the input text and fusing and aligning labeled subtokens. The predictions of all classifiers are then wrapped in a common result object that implements the different output formats shown in Listing~\ref{lst:example}. This design means that adding a new model to \texttt{pylazaro} only requires implementing a new classifier that loads the model and returns BIO tags, while the rest of the library (and the code of its users) remains unchanged.

\section{Demo}\label{sec:demo}

\begin{figure*}[t]
  \includegraphics[width=\textwidth]{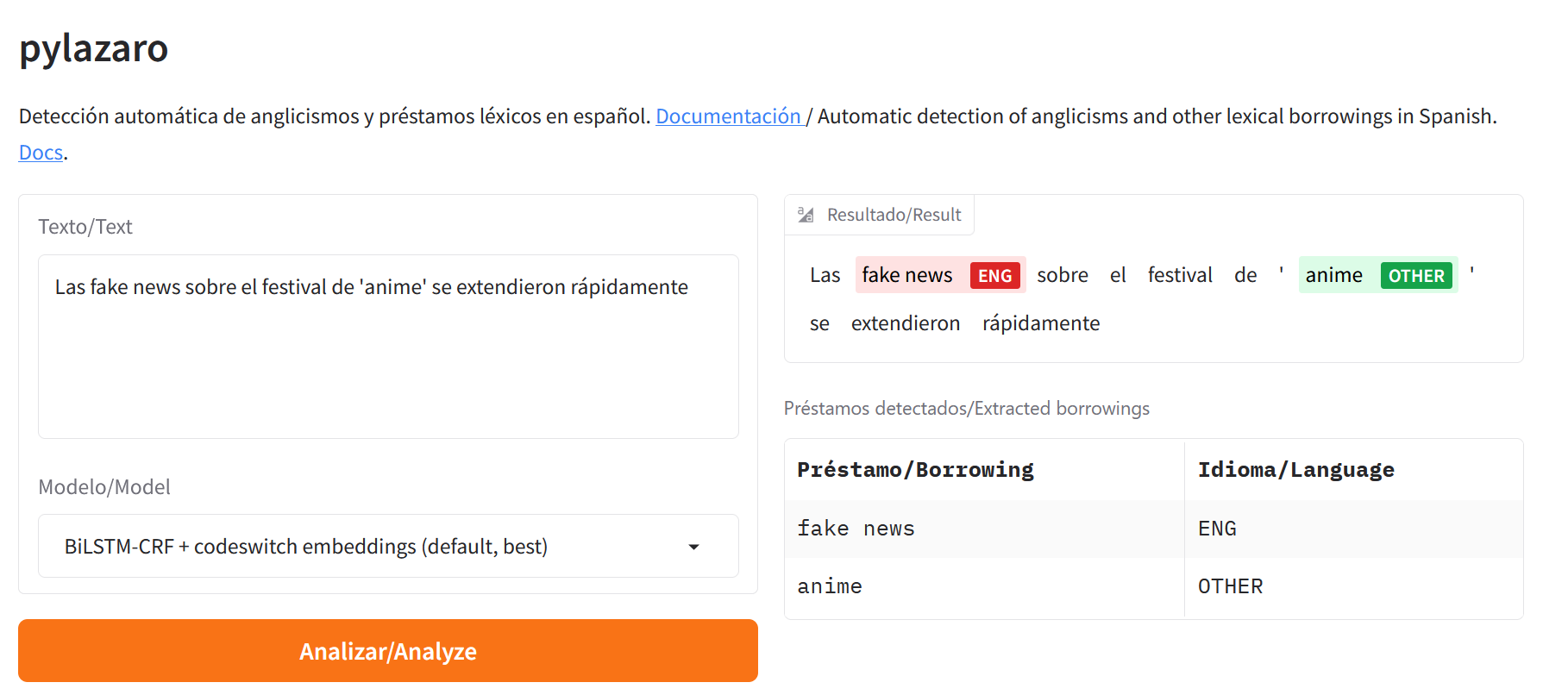}
  \caption{Screenshot of the \texttt{pylazaro} demo hosted on HuggingFace Spaces. The borrowings detected in the input text are highlighted according to their label (\texttt{ENG} or \texttt{OTHER}) and listed in a table.}
  \label{fig:demo}
\end{figure*}

In order to make \texttt{pylazaro} accessible to users without a programming background (such as lexicographers or linguists), we provide a live demo of the library hosted on HuggingFace Spaces\footnote{\url{https://huggingface.co/spaces/lirondos/pylazaro-demo}} (see Figure~\ref{fig:demo}). The demo was built with Gradio and runs \texttt{pylazaro} under the hood. The user can type or paste a text in Spanish and select one of the neural models available in the library. The demo then displays the input text with the retrieved borrowings highlighted according to their label, together with a table that lists every borrowing found in the text and its language. A set of example sentences is provided so that users can try the tool without having to come up with their own text.

Because all models are exposed through the same interface, the demo also makes it easy to compare how different models behave on the same input, for instance when dealing with multiword borrowings or with borrowings that appear in different positions within the sentence \citep{alvarez2024characterizing}.

\section{Comparison against LLMs}
\label{sec:llm}
All the models behind \texttt{pylazaro} are medium-sized models fine-tuned specifically for the task of retrieving anglicisms from Spanish. One could reasonably argue that the task of automatically extracting words of English origin from Spanish text can simply be achieved by using any available general-purpose LLM. 

Prior work, however, has shown that LLMs are not good at this task: a collection of 23 autoregressive LLMs (including models from the Qwen, Gemma and Llama3.1 families) were tested for the task of extracting anglicisms from Spanish text \citep{GONZALEZ2026101899}. The results ranged from 0.0 to 0.32 of F1 score, with Microsoft Phi-4, 40B-ALIA and the 9B-EuroLLM ranking in the first positions. The twenty remaining models all scored below 0.2 of F1 score. 
Similarly, \citet{alvarez-mellado_lexical_2025} reported that 8B-Llama3 obtained an F1 score of 0.39 on the same evaluation set (with a different prompting strategy).

These poor results contrast with the scores obtained by the finetuned models such as the ones behind \texttt{pylazaro}, whose scores ranged between 0.83 and 0.86 of F1 scores over the same evaluation set. 

\begin{table}[t]
\centering
\small
\begin{tabular}{lrr}
\toprule
Model & Params & F1 \\
\midrule
BiLSTM-CRF (codeswitch) & 182M & 0.86 \\
BiLSTM-CRF (BETO/BERT) & 226M & 0.84 \\
mBERT & 177M & 0.84 \\
BETO & 109M & 0.83 \\
\midrule
Llama3 & 8B & 0.39\\
Phi-4 & 14B & 0.32 \\
ALIA-instruct & 40B & 0.28 \\
EuroLLM-instruct & 9B & 0.26 \\
Remaining 20 LLMs & -- & $<$0.20 \\
\bottomrule
\end{tabular}
\caption{F1 scores of the neural models behind \texttt{pylazaro} \citep{alvarez-mellado_detecting_2022} and of the best performing general-purpose LLMs \citep{alvarez-mellado_lexical_2025,GONZALEZ2026101899} for the task of anglicism extraction in Spanish.}
\label{tab:llms}
\end{table}

In addition to their better performance, the models behind \texttt{pylazaro} are considerably smaller than general-purpose LLMs: while the models in \texttt{pylazaro} have between 109 and 226 million parameters, the best performing LLMs have between 9 and 40 billion parameters. As a consequence, \texttt{pylazaro} can be run on a regular laptop without a GPU, which is a relevant factor for linguists who need to process large collections of text.

This justifies having a dedicated library for anglicism extraction in Spanish that can perform better and more efficiently than general purpose LLMs and that can easily be adopted by experts for their linguistic tasks.

\section{Adoption and use cases}
\label{sec:adoption}
According to \textit{pepy.tech}\footnote{\url{https://pepy.tech/projects/pylazaro}}, as of September 2026, \texttt{pylazaro} has been downloaded more than 58,000 times from PyPI, with over 800 downloads over the last 30 days. 
The models behind \texttt{pylazaro} have also been downloaded over  63,000 times from the HuggingFace Hub.
\texttt{pylazaro} is also the library behind \textit{Observatorio Lázaro}\footnote{\url{https://observatoriolazaro.es/}} \citep{alvarezmellado2026observatoriolazaroselfpopulatingdatabase}, a pipeline that monitors anglicism usage in the Spanish press. The site, which has collected over 2 million anglicisms since 2020, showcases the type of corpus linguistic analysis that \texttt{pylazaro} can facilitate.
Other potential uses of this library include  preprocessing of corpora to assist historical linguists track language change in text or help lexicographers identify words that are candidate to be registered in dictionaries.  

\section{Conclusions}
In this paper we have introduced \texttt{pylazaro}, a Python package that identifies unassimilated lexical borrowings (mostly anglicisms) in Spanish text. \texttt{pylazaro} offers a single interface to five models that rely on different libraries, and it can be used either as a library or through a live demo that requires no installation. To the best of our knowledge, \texttt{pylazaro} is the first available library for the automatic identification of borrowings in running text. The poor performance of general-purpose LLMs at this task and the adoption of the library, which has been downloaded more than 58,000 times and powers \emph{Observatorio L\'azaro}, show the need for dedicated tools for borrowing detection.

\section*{Limitations}

The models behind \texttt{pylazaro} were trained on an annotated corpus of European Spanish newspaper text \citep{alvarez-mellado_detecting_2022}. Their performance on other domains (such as social media) or on other varieties of Spanish has not been systematically evaluated and is likely to be lower. Additionally, \texttt{pylazaro} only retrieves unassimilated borrowings: borrowings that have already been adapted to Spanish orthography or morphology are not detected. Although the tagger distinguishes between anglicisms and borrowings from other languages, the models have mostly been optimized for anglicisms, and borrowings from other languages are underrepresented in the training data. Finally, because \texttt{pylazaro} relies on external deep learning libraries, keeping the models compatible with new releases of those libraries requires ongoing maintenance.


\bibliography{custom,references}

\end{document}